\documentclass[UTF8,10pt,journal,cspaper,compsoc]{IEEEtran}
\usepackage{cite}

\ifCLASSINFOpdf
  \usepackage[pdftex]{graphicx}
  \DeclareGraphicsExtensions{.pdf,.jpeg,.png}
\else
\fi

\usepackage[cmex10]{amsmath}
\usepackage{amssymb}

\usepackage{algpseudocode}

 \usepackage[utf8]{inputenc}
\newtheorem{definition}{Definition}

\begin{document}
\markboth{ACCEPTED TO APPEAR IN IEEE TRANSACTIONS ON SERVICES COMPUTING 2015, DOI  10.1109/TSC.2015.2402679}{}

%
\title{An Integrated Semantic Web Service Discovery and Composition Framework}

\author{Pablo~Rodriguez-Mier,
        Carlos~Pedrinaci,
	Manuel~Lama,
	and~Manuel~Mucientes
\IEEEcompsocitemizethanks{\IEEEcompsocthanksitem P. Rodriguez-Mier, M. Lama and M. Mucientes
work at the Centro de Investigación en Tecnoloxías da Información (CiTIUS), 
Universidade de Santiago de Compostela, Spain.\protect\\
E-mail: \{pablo.rodriguez.mier,manuel.lama,manuel.mucientes\}@usc.es
\IEEEcompsocthanksitem Carlos Pedrinaci is with The Open University, Milton Keynes, UK.\protect\\
E-mail: carlos.pedrinaci@open.ac.uk}

\thanks{
  © 2015 IEEE. Personal use of this material is permitted. Permission
  from IEEE must be obtained for all other uses, in any current or future media, including
  reprinting/republishing this material for advertising or promotional purposes,
  creating new collective works, for resale or redistribution to servers or lists, or reuse of any
  copyrighted component of this work in other works.
}}


\IEEEcompsoctitleabstractindextext{%
\begin{abstract}

In this paper we present a theoretical analysis of graph-based service composition in terms of its dependency with service discovery. Driven by this analysis we define a composition framework by means of integration with fine-grained I/O service discovery that enables the generation of a graph-based composition which contains the set of services that are semantically relevant for an input-output request. The proposed framework also includes an optimal composition search algorithm to extract the best composition from the graph minimising the length and the number of services, and different graph optimisations to improve the scalability of the system. A practical implementation used for the empirical analysis is also provided. This analysis proves the scalability and flexibility of our proposal and provides insights on how integrated composition systems can be designed in order to achieve good performance in real scenarios for the Web.
\end{abstract} 

}

\maketitle

\IEEEdisplaynotcompsoctitleabstractindextext

%
\IEEEpeerreviewmaketitle

\section{Introduction}
\label{Section:Introduction}

\IEEEPARstart{S}{ervice} discovery and composition are in general complex tasks that require considerable effort, especially when vast amounts of services are available. Service discovery solutions range from the initial UDDI proposal 
that relied on the syntactic description of services and a prefixed categorisation~\cite{alonso2004web}, to more advanced generic solutions able to discover Web APIs and Web services across domains exploiting rich user-provided semantic service descriptions~\cite{Pedrinaci2010}. Similarly, a plethora of service composition solutions have been produced spanning from mere graphical support to completely automated solutions~\cite{Srivastava2003,rao2005survey,Dustdar2005Survey}. Both discovery and composition engines essentially rely on the processing of service descriptions, which increasingly go beyond syntactic representations to include the semantics of the 
service(s) 
to enable more advanced computations~\cite{Mcilraith,Papazoglou:2007p162}.

An analysis of the service composition literature highlights that, regardless of the approach, a central task that needs to be frequently performed throughout the composition activity, is the discovery of suitable services to use. Whether one looks at fully automated composition engines based on Artificial Intelligence (AI) planning techniques~\cite{Bertoli2003,Sirin2004,Klusch2005}, or at more constrained solutions that rely on pre-defined skeletal plans~\cite{McIlraithSheila2002, Sirin2005template}, or at
graph based approaches focused on semantic input-output parameter matching \cite{Kona2008,Yan2008,MarcoAiello2008,Nam2008a,Raman2008,Shiaa2008,Hennig2010,Oh2007web},
service discovery is a central activity that needs to be carried out at every main step during the generation of the composition. Yet, despite the strong dependency between both activities, research and development in both areas has evolved for the most part independently.

On the one hand, service discovery has traditionally been approached as a one-of activity to be sporadically carried out by humans when looking for services. 
As a consequence the interface exposed by discovery engines assumes that requests are fully specified in terms of a well-defined interface and categorisation. 
Moreover, response times of discovery engines are orders of magnitude above what would be acceptable for a composition engine that should it delegate the thousands 
discovery requests it needs to issue at composition time \cite{klusch2012overview}. These limitations hamper the development of fast composition systems where discovery and composition
are two fundamental, interrelated activities. 

On the other hand, partly due to the particularly demanding computational needs of composition algorithms, most composition engines reimplement locally their own discovery methods instead of integrating existing components providing state of the art discovery algorithms. Additionally, this approach relies on the unnecessary and often unrealistic assumption that the entire set of services should be locally available to the composition engine. This assumption requires pre-importing all services locally which is only viable for those registries providing entire public dumps of the service descriptions they hold. Furthermore, most composition engines do not introduce optimisation techniques to improve the scalability by identifying equivalent or dominant functionality that could appear when many differents service registries are involved in the composition. This prevents the use of optimal search strategies since the complexity usually grows exponentially with the number of services.

In order to tackle the previous problems, a composition framework should consider the following characteristics:
1) provide convenient \textit{fine-grained discovery} mechanisms that could help to discover services able to consume or produce (a subset of) certain types of data as usually required during composition; 2) improve the \textit{response time} of service discovery to process requests very fast; 3) support the integration of \textit{third
party service registries} as a key activity in the composition phase; 4) incorporate \textit{optimizations} to improve the scalability of the overall composition
process; and 5) find \textit{optimal service compositions} by minimizing different criteria such as the number of services or the length of the composition to avoid complex, unmanageable solutions.

In this paper we present a graph-based framework focused on the semantic input-output parameter matching of services' interfaces that
efficiently integrates the automatic service composition and semantic service discovery. The provided framework takes into
account all the characteristics indicated in the above paragraph. Notably, the main contributions are:

\begin{IEEEenumerate}
 \item A formal framework that presents a theoretical analysis of graph-based service composition
in terms of its dependency with a service discovery and we provide a fine-grained I/O discovery interface
which reduces the performance overhead without having to assume the local availability and in-memory preloading of service registries. 
The framework also includes an optimal composition search algorithm to extract the best composition from the graph minimising the length and the number of services, and different graph optimisations to improve the scalability of the system, which as far as we now are not included in other frameworks. 
 \item A reference implementation of this formal framework based on the adaptation of two independently developed components, namely ComposIT~\cite{RodriguezMier2012} and iServe~\cite{Pedrinaci2010}, respectively in charge of service composition and discovery.
 \item A detailed performance analysis of the integrated system, highlighting both the unacceptable performance achieved when using the typical out of the box discovery implementations, as well as the fact that top performance is achievable with the adequate discovery granularity and corresponding indexing optimisations.
\end{IEEEenumerate}

The proposed framework is data-flow centric, focused on the semantic I/O parameter matching of services' interfaces and leaving aside preconditions and effects. This is essentially a pragmatic decision inline with the current tendency towards lightweight data-driven approaches. In fact, on the Web less than 5\% of the semantic Web services include 
preconditions and effects~\cite{Pedrinaci:2010e}. 

The rest of the paper is organized as follows. Sec. \ref{Section:RelatedWork} discusses the state-of-the-art.  
Sec. \ref{Section:WSCProblem} formalizes the web service composition problem and 
Sec. \ref{Section:Framework} framework that defines the composition in terms of service discovery tasks. 
Sec. \ref{Section:Architecture} describes our reference implementation. 
Sec. \ref{Section:Evaluation} explores the performance of the system for different scenarios and finally 
Sec. \ref{Section:Conclusions} gives some final remarks. 

\section{Related Work}
\label{Section:RelatedWork}

Automatic composition of Web services is still an open problem that involves multiple
research areas \cite{Dustdar2005Survey}. Concretely, lots of efforts have been devoted 
to automate the discovery and composition using different approaches and techniques \cite{pedrinaci2011semantic}.
However, most of the research in both areas has been evolved independently of each other, despite the
significant overlap between these interrelated tasks. This has lead to a lack of integrated approaches in the field
that consider the performance and the scalability of the overall integrated system as well as the impact of
the discovery in terms of response time during the automatic composition task.

From the discovery side, most of the work has been focused on improving the retrieval performance (i.e., precision-recall curve) without
much concern about the response time requirements and/or the interface requirements to provide an efficient fine-grained discovery granularity 
for automatic composition. However, the response time of the discovery systems is recently gaining significant interest. 
A recent service discovery competition \cite{klusch2012overview} shows some of the newest advances in the automatic
discovery field. Most relevant examples are OWLS-MX3\cite{klusch2009owls}, iSem 1.1\cite{klusch2010isem} and XSSD\cite{chu2011xssd}. The main
conclusions that can be drawn from this contest, from the perspective of service composition, are twofold: 1) research efforts are focused on response 
time improvement via caching and indexing, yet still not sufficient 
for fast, automatic composition of services and 2) the interface exposed by discovery engines
assumes that requests are fully specified in terms of a well-defined interface and categorisation, i.e., 
discovery systems expect a precise description of the service in terms of inputs and outputs, and/or other characteristics such as preconditions and effects. However, these interfaces are not adequate for service composition, since one of the assumptions is that there is usually no single service that fully matches a request and therefore several services need to be combined instead. Indeed, during automatic composition, an exploratory search is usually required to guess which relevant services can be selected at each step. This requires to launch many partial requests (fine-grained queries), rather than fully specified requests, in order to locate relevant services that match some partial information available to the algorithm (e.g., services that consume some inputs and/or produce some outputs). Fine-grained requests are simpler and can be solved faster than complex, fully specified requests. Thus they are more suitable for automatic composition systems.

From the composition side, most approaches can be categorized into: 1) classical AI planning approaches\cite{Peer2005}, where the composition
problem is translated into the planning domain and solved using general planners, and 2) graph-based I/O driven approaches that build a graph
with the services and their input/output semantic relations (generally ommiting the preconditions and effects),
and apply graph search techniques to extract (usually optimal) service compositions from the graph.

Relevant approaches of the first group are \cite{Sirin2004,Klusch2005,Hatzi2011}. 
These approaches differ from our work in the sense that they handle very expressive preconditions and effects to generate composition plans
but: 1) the concept of external service registries is missing, services are assumed to be locally available; 2) average response time of these systems is usually high; 
and 3) optimizations to reduce the number of services by identifying redundant functionality are not considered.

On the other hand, graph-based I/O approaches are gaining much attention since the Web Service Challenge~\cite{Bansal2008}.
Some notable works in this field are \cite{Yan2008,MarcoAiello2008,Nam2008a,Raman2008,Shiaa2008,Hennig2010,Oh2007web}. 
Concretely, \cite{Yan2008,MarcoAiello2008,Oh2007web} are the top-3 algorithms of the WSC'08. Although these approaches
show generally good performance and low response times, \cite{Yan2008} and \cite{MarcoAiello2008} do not find optimal solutions and \cite{Oh2007web}
fails to find solutions in large data sets. Additionally, none of these systems consider neither the integration with service registries nor the use
of service optimizations to deal with potential scalability problems.

From the point of view of the integrated frameworks, a very interesting approach was proposed by Kona et al. in \cite{Kona2008}. In this paper,
the authors present an efficient framework for Web service composition that supports semantic Web service discovery.
The composition is generated by performing a forward chaining of operators to find a feasible composition. 
The authors also evaluated the system with the datasets of the Web Service Challenge 2006 and presented a detailed experimentation. 
Their results demonstrate the capabilities and the good performance of this system which, however, exhibits some limtations:
1) the notion of an external service registry is missing, all the information required is preprocessed and loaded 
in the main memory, which is one of the main issues we set out to tackle with this work since it is otherwise not possible to deal with large and/or distributed datasets; 2) the framework does not contemplate
service optimisations to remove redundant information and 
3) the work does not perform an optimal search to minimise the cost or the number of services of the composition as all possible compositions
with the shortest length are captured in the composition graph which should be further processed to extract the optimal composition.
Similarly, in \cite{Lecue2007}, Lécué et al. develop an integrated framework for dynamic Web service composition. 
The framework exploits the semantic input-output matchmaking to discover relevant services and performs automatic composition using a graph-based
approach, taking into account functional and non-functional properties. However, graph optimisations are not considered and the composition search is non-optimal,
since the selection of the services is merely greedy-based. 

In \cite{DaSilva2011}, Da Silva et al. present a framework that effectively supports both automatic semantic discovery and composition, among other
relevant phases of the composition life-cycle, such as service publication and service selection, taking into account non-functional properties.
One of the limitations of the discovery phase is that it does not support fine-grained requests. 
On the other hand, the framework does not include neither optimisations 
to reduce graph size nor an optimal search to extract the best composition from the graph.

In light of the above analysis, we propose a graph-based I/O framework that overcomes all of the analyzed limitations.
In this framework the discovery is defined in terms of a fine-grained I/O interface which minimises
the performance overhead between both composition and discovery without having to assume the local availability and in-memory preloading of service registries.
The proposed framework also includes an optimal composition search 
algorithm to extract the best composition from the graph minimising the length and the number of services,
and different graph optimisations to improve the scalability of the system.

\section{Web Service Composition Problem}
\label{Section:WSCProblem}

Service composition aims to help construct composite services that could fulfil a user request, e.g., booking an entire holiday, when no known service can achieve such a request on its own. A core activity for creating service compositions is, indeed, the discovery of relevant services. In this context, relevant services are those that could be invoked and contribute to obtaining an executable composition that would fulfil the needs set out by the client. We herein formalise the composition problem in close relationship with discovery as a means to better study and approach the integration of discovery and composition engines. The formalisation of the problem is data-flow centric, focussed on the semantic input-output parameter matching of services' interfaces.

\subsection{Semantic Web Service Discovery}
The semantic Web service discovery problem consists of locating appropriate services 
from one or more service registries that are relevant to an input-output request.

\begin{definition}
 A Semantic Web Service (SWS, hereafter ``service'') can be defined as a 
 tuple $w=\{In_w, Out_w\} \in W$ where $In_w$ is a set of inputs 
 required to invoke $w$, $Out_w$ is the set of outputs returned by $w$ after its execution, and $W$ is
 the set of all services available in the service registry. Each input and output is related to a semantic concept from
 an ontology $O$ ($In_w,Out_w \subseteq O$).
\end{definition}

Semantic inputs and outputs can be used to discover relevant services as well as to compose the functionality
of multiple services by matching their inputs and outputs together. In order to measure the quality of the match, we need a matchmaking 
mechanism that exploits the semantic I/O information of the services. The different matchmaking degrees that are 
typically contemplated in the literature are \cite{Paolucci2002}:

\begin{itemize}
 \item \textbf{Exact ($\equiv$)}: An output $o_{w1} \in Out_{w1}$ of a service $w1$ matches an input $i_{w2} \in In_{w2}$
  of a service $w2$ with a degree of exact match if both
  concepts are equivalent.
 \item \textbf{Plugin ($\sqsubseteq$)}: An output $o_{w1} \in Out_{w1}$ of a service $w1$ matches an input $i_{w2} \in In_{w2}$ of a service $w2$
 with a degree of plugin if $o_{w1}$ is a sub-concept of $i_{w2}$ ($o_{w1} \sqsubseteq i_{w2}$).
 \item \textbf{Subsume ($\sqsupseteq$)}: An output $o_{w1} \in Out_{w1}$ of a service $w1$ matches an input $i_{w2} \in In_{w2}$ of a service $w2$
 with a degree of subsume if $o_{w1}$ is a super-concept of $i_{w2}$ ($o_{w1} \sqsupseteq i_{w2}$).
 \item \textbf{Fail ($\perp$)}: When none of the previous matches are found, then both concepts are 
 incompatible and the match has a degree of fail ($o_{w1} \perp i_{w2}$).
\end{itemize}

Note that, in order to discover relevant services to generate data-flow compatible service compositions, 
the only two valid degrees of match are \textit{exact}
and \textit{plugin}. On this basis, we define
the \textit{cmatch} (compatible match) function that will be used to discover candidate services during the composition phase:

\begin{definition}
\label{def:cmatch}
 Given $a,b \in O$, a compatible match \textit{cmatch(a,b)} 
 holds if and only if $a \equiv b$ (exact match) or $a \sqsubseteq b$ (plug-in match).
\end{definition}

Using the previous compatible match function between concepts, we can define the matchmaking operator ``$\otimes$'' that given two
sets of concepts $C_1, C_2 \subseteq O$, it returns the concepts from $C_2$ matched by $C_1$.

\begin{definition}
\label{def:matchmaking}
 Given $C_1, C_2 \subseteq O$, we define ``$\otimes: O \times O \rightarrow O$'' such that
 $C_1 \otimes C_2 = \{c_2 \in C_2 | cmatch(c_1, c_2), c_1 \in C_1\}$. Note that this operator
 is not commutative.
\end{definition}

We can use the previous operator to define the concepts of full and partial matching between
concepts.

\begin{definition}
 Given $C_1, C_2 \subseteq O$, a full matching between $C_1$ and $C_2$ exists if $C_1 \otimes C_2 = C_2$, whereas
 a partial matching exists if $C_1 \otimes C_2 \subset C_2$.
\end{definition}

Typically, a service $w=\{In_w, Out_w\}$ is relevant to a request $r=\{In_r, Out_r\}$, where $In_r \subseteq O$ 
are the provided inputs and $Out_r \subseteq O$ the expected outputs,
if $In_r \otimes In_w = In_w$ and $Out_w \otimes Out_r = Out_r$, that is, there is a full match between the provided inputs and the service inputs and a full match between the service outputs and the expected outputs. 

While this approach is reasonable for discovering the services that best match an entire request (full match), for composition one needs to locate services that are relevant, that is, that match some inputs / outputs (partial match). Thus, rather than approaching the discovery problem based on a full input/output description, we split this problem into two finer-grained discovery problems that are more relevant for service composition: input discovery and output discovery.

\begin{definition}
\label{def:inputrelevant}
 Given a set of concepts $C \subseteq O$, the input discovery problem can be defined as finding a set of relevant services 
 $W=\{w_1,...,w_n\}$ where $w_i=\{In_{w_i}, Out_{w_i}\}$ such that $\forall w_i \in W$, $C \otimes In_{w_i}  \subseteq In_{w_i} $, that is,
 services that can consume some (partial match) of the inputs or are directly invokable (full match) with $C$.
\end{definition}

\begin{definition}
\label{def:outputrelevant}
 Given a set of concepts $C \subseteq O$, the output discovery problem can be defined as finding a set of relevant services 
 $W=\{w_1,...,w_n\}$ where $w_i=\{In_{w_i}, Out_{w_i}\}$ such that $\forall w_i \in W$, $Out_{w_i} \otimes C \subseteq C$, that is,
 services that produce some or all outputs.
\end{definition}

Based on these definitions, we introduce the notion of input and output relevance:

\begin{definition}
\label{def:relevance}
 A service $w=\{In_w, Out_w\}$, where $In_w, Out_w \subseteq O$,  is \textit{input-relevant} for a set of concepts $C \subseteq O$ if
 $C \otimes In_w \neq \emptyset$, whereas the service $w$, is \textit{output-relevant} for a set of concepts $C \subseteq O$ if
 $Out_w \otimes C \neq \emptyset$.
\end{definition}


\begin{figure*}[htpb]
  \centering
  \includegraphics[width=0.80\textwidth,height=\textheight,keepaspectratio]{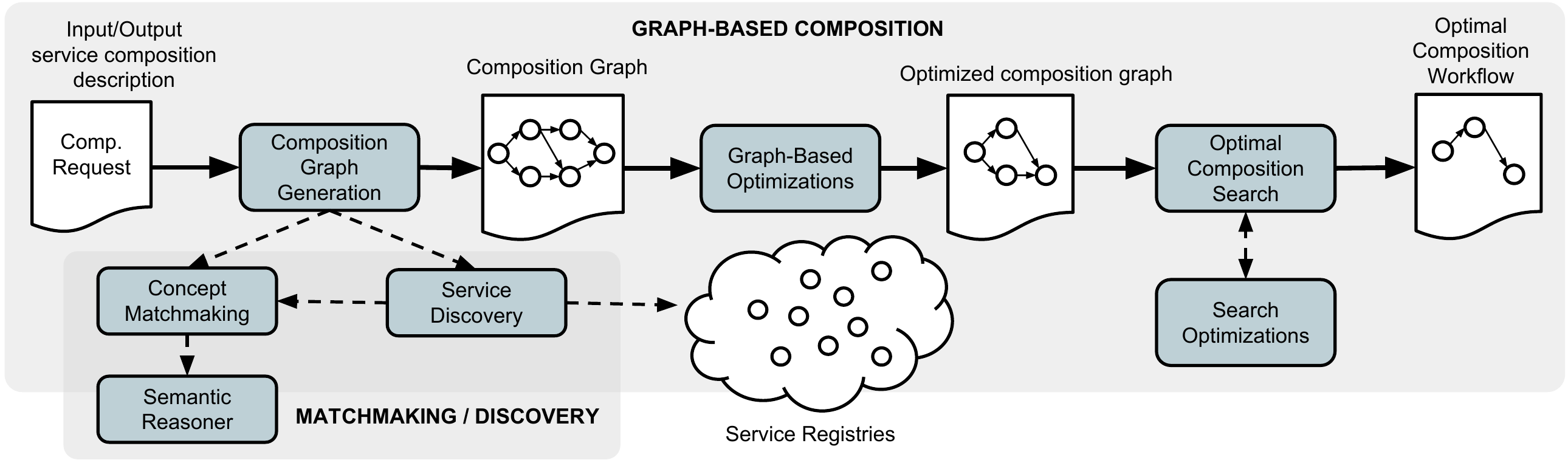}
  \caption{Overview of the proposed approach.}
  \label{Figure:Overview}
\end{figure*}

\subsection{Semantic Web Service Composition}

The semantic composition problem considered in this work is as follows: 
Given a request $r=\{In_r, Out_r\}$, where $In_r$ is a set of available semantic input concepts and $Out_r$ a set of requested semantic output concepts,
we can define the problem of the automatic construction of a SWS composition as that of finding a composite Web service
$w_c=\{In_{w_c}, Out_{w_c}, P=\{S,\leq\}\}$ such that $In_r \otimes In_{w_c} = In_{w_c}$ (the composite service is invokable with the available inputs)
and $Out_{w_c} \otimes Out_r = Out_r$ (the composite service retrieves all the requested outputs). This service consists of a partially
ordered set $P$ (a binary relation ``$\leq$'' over a set of services $S \subseteq W$). This partial ordered set of services is esentially a 
Directed Acyclic Graph (DAG) which models the implicit execution order of the services driven by the input/output matches, 
where nodes of the DAG are services and the arcs are valid semantic matches. 
This type of composition has many advantages: On one hand, mapping inputs and outputs to semantic concepts does allow to reason about
data types to improve the matchmaking between service parameters, which leads to more possible semantically valid compositions.
On the other hand, DAG representation formally captures 
the nature of a composition where services may be executed in different orders, i.e., there are many different total (sequential) orderings of a composition
that lead to the same result. Moreover, since our approach is data-flow centric, a DAG representation is simpler than a general (possible cyclic) graph 
as cycles do not produce new data types in the composition. 

However there are also some drawbacks. First, a DAG representation could impose some restrictions in the compositions that can
be generated, i.e, due the absence of cycles, a service could not explicitly be invoked twice. Second, compositions at different semantic levels
rather than just concept matchmaking would deffinitely improve the quality of the compositions by capturing more possible cases.
Furthermore, using input concepts and output concepts to define a composition request is not user friendly.  A better way to specify
a request would be to define it with keywords. This, nonetheless, could be achieved with a pre-processing step using 
automatic semantic annotation tools to translate the request from keywords to semantic concepts.
Formally, we define a valid composition as follows:

\begin{definition}
\label{def:composition}
Let $r=\{In_r, Out_r\}$ and let $w_c=\{In_{w_c}, Out_{w_c}, P=\{S,\leq\}\}$ be a composite service for the request $r$, where $P$ is a partial order
over the set of services $S \subseteq W$ of the composite service $w_c$. We say $w_c$ is a valid composition for request $r$ if and only if, for any
topological sort $T=\{w_1,w_2,...,w_N\}$ of $P$, where $w_j=\{In_{w_j},Out_{w_j}\}$ $\forall j \in [1,N] $, the following expression is
satisfied:
\begin{eqnarray*}
 (In_r \otimes In_{w_1} = In_{w_1}) \wedge ((In_r \cup Out_{w_1}) \otimes In_{w_2} = In_{w_2}) \nonumber \\
 \wedge \ldots \wedge ((In_r \cup Out_{w_1} \cup ... \cup Out_{w_N}) \otimes Out_r = Out_r).
\end{eqnarray*}
\end{definition}

This definition implies that every service of the composition must be invokable to obtain an invokable service composition.
We say that a service $w=\{In_w,Out_w\}$ is invokable with a set of concepts $C \subseteq O$ if each required input 
$i_w \in In_w$ is semantically matched by a set of concepts $C$.

\begin{definition}
  If $C \subseteq O$ is the set of available input concepts, then a service $w=\{In_w, Out_w\}$ 
  is \textit{invokable} with $C$ if $C \otimes In_w = In_w$, i.e., there exists a full matching between the available inputs and the service inputs.
\end{definition}

Note that if a service $w$ is invokable with a set of concepts $C$, then it is also input-relevant for the 
same set of concepts since \textit{invokable} implies \textit{input-relevant}, but the inverse does not hold (see Def. \ref{def:relevance}).
That is, the set of invokable services is included in the set of the relevant services. 

The reader should note that we restrict the definition of a compatible match to \textit{exact} and \textit{plugin} in order to generate semantically complete compositions.
However, the framework also supports the use of other match degrees (e.g., subsume) by relaxing the ``$cmatch$'' operator, which in practice means obtaining potentially more matched
(but semantically weaker) concepts and thus bigger composition graphs with more services and match relations that could be semantically incomplete. 
This is supported not only in theory, but also by the reference implementation presented in Sec. \ref{Section:Architecture}.

\section{Composition Framework} 
\label{Section:Framework}

On the basis of the formal definition of the problem, 
in this section we present a graph-based framework for automatic semantic Web service composition. 
Fig. \ref{Figure:Overview} shows the overview of our approach with the different steps involved. The process is triggered by a composition request that specifies the user requirements in terms of inputs and the expected outputs. 
This information is used in the composition graph generation phase to build a graph with all the relevant services and the
semantic relations between their inputs and outputs. In order to find the relevant services, the composition graph phase is
interleaved with the \textit{discovery phase}. The discovery phase is responsible for retrieving the relevant services given the data available at different stages during the \textit{composition graph generation} phase. The relationships between the inputs and outputs of services are computed
in the \textit{matchmaking phase}, where the semantic matching degree between inputs and outputs is computed using a semantic reasoner. The service composition graph is eventually generated on the basis of the relevant services and the I/O matching information. This graph contains all possible service compositions that satisfy the composition request, in addition to a few others that, although invokable, do not manage to entirely fulfil the request. The service composition graph is then optimised applying different techniques to group and reduce the number of services and relations. Next, an \textit{optimal search} is performed over the graph to find the optimal composition. This phase is interleaved with a \textit{search optimisation phase} that analyses and reduces the search space. Finally, the optimised composition workflow is returned. 

In this section, we analyse each phase and we provide generic strategies based on the problem description presented in the previous section.

\subsection{Semantic Matchmaking}
A fundamental functionality that needs to be available for generating compositions and even for discovering services, is the ability to analyse the compatibility between different semantic types. This functionality, which we refer to as semantic matchmaking, is in charge of assessing the level of semantic compatibility between concepts, given an ontology (or set of ontologies). To do so, semantic matchmaking relies on semantic reasoning (notably subsumption reasoning) in order to be able to determine the relationships between the concepts (e.g., Plugin match). This mechanism can be used for example, to discover services that can consume or produce a concrete input/output by finding semantically compatible types. Such a mechanism is also particularly relevant for generating the service composition graph with all the matches between services inputs and outputs.

The matchmaking system provides a $match(C_1, C_2)$ function which represents the concrete implementation function of the $\otimes$ operator
defined in Def. \ref{def:matchmaking}. The $match$ function tries to find a valid match between the \textit{source concepts} of $C_1$
and the \textit{target concepts} of $C_2$ calling the $cmatch(c_i, c_j)$ function (Def. \ref{def:cmatch}) for each pair $(c_i,c_j)$
of concepts where $c_i \in C_1$ and $c_j \in C_2$. The compatible match
function is calculated using a semantic reasoner that returns the semantic relation between two concepts. Then, it checks
if the relation is considered a compatible match (i.e., \textit{exact} or \textit{plugin}).
Each time a compatible match is found between $c_i$ and $c_j$, $c_j$ is added to a set of matched concepts and removed from $C_2$. 
The reader should note that the goal here is not to find the best match for each element but rather to get all compatible matches for each target element.

The best-case complexity (all $C_2$ concepts matched by the first element from $C_1$) is $O(m)$, whereas
the worst-case complexity (no compatible matches at all) is $O(m \cdot n)$ where $n=|C_1|, m=|C_2|$. This implies
that, in the worst case, for two sets of elements, there will be at most $m \times n$ 
calls to the \textit{cmatch} function which is ultimately answered by the semantic reasoner. 

\subsection{Semantic Service Discovery}
\label{Subsection:Discovery}
In order to generate service compositions, it is necessary to be able to discover appropriate services based on their interface. The goal of a typical discovery system is to find atomic services that match entirely a description representing the ideal service sought, i.e., all the inputs and outputs are compatible. However, from the viewpoint of generating data-flow compatible compositions, rather than looking for entire matches, we need to find suitable combinations of services that combined would satisfy a request. In this scenario, the ability to find partially matching services very fast is paramount in order to enable exploring efficiently the many possible combinations of services that could lead to a suitable composition. 
Therefore, in a nutshell, the type of service discovery that is required for supporting service composition is a more relaxed and finer-grain version of that typically provided by discovery engines whereby partial matches can be obtained in a very fast manner. This can be achieved by defining a simple fine-grained interface that supports the discovery of services using only partial information (some/any available inputs, some/any expected outputs). Fig. \ref{Alg:RelevantIO} shows the pseudocode of this simple interface to discover relevant services that can be used as a starting point to obtain semantic input/output relevant services, as defined in Def. \ref{def:relevance} in Sec. \ref{Section:WSCProblem}. 
 
The discovery algorithm sequentially scans all services and calls the $Match$ function
of the \textit{Matchmaker} to determine if a service is relevant for an input (the service has at least one input compatible with the inputs
provided) or for an output (the service has at least one output compatible with the outputs provided) depending on the $Type$ selected.
Therefore, the complexity of this type of discovery is $O(w)$ where $w=|W|$ is the size of the service repository. This implies at most $|W|$ calls to
$Match$ in the worst-case scenario or $O(w \cdot m \cdot n)$ if we consider the complexity of the $Match$ method
assuming every service has at most $m$ outputs and $n$ inputs.

\begin{figure}
{\footnotesize
\begin{algorithmic}[1]
\Function{RelevantIO}{$C \subseteq O, W, type$}    
  \State $relevantServ := \{\}$
  \ForAll{$w_i=\{I_{w_i},O_{w_i}\} \in W$}
      \If{$type = In$}
	\If{$match(C, I_{w_i})$}
	  \State $relevantServ := relevantServ \cup w_i$
	\EndIf
      \ElsIf{$type = Out$} 
	\If{$match(O_{w_i}, C)$}
	  \State $relevantServ := relevantServ \cup w_i$
	\EndIf
      \EndIf
  \EndFor
  \State \Return $relevantServ$
\EndFunction
\end{algorithmic}}
\caption{Pseudocode to obtain input-relevant and output-relevant sets of services for a particular set of concepts}
\label{Alg:RelevantIO}
\end{figure}

\subsection{Service Composition Graph Generation}
When the system receives a request, the \textit{Service Composition Graph Generator} computes a graph with all the semantic relations
between the relevant services for the request. A request is basically a set of input concepts, which represent the initial set of available inputs, and a set of output concepts, which
are the outputs that the composite service should return. 
The \textit{service composition graph} is basically a layered Directed Acyclic Graph (DAG), $G=(V,E)$, where:
\begin{itemize}
 \item $V = W \cup C$ is the set of vertices of the graph, where $W$ is the set of services and $C$ the set of concepts
 (inputs and outputs).
 \item $E = CW \cup WC \cup CC$ is the set of edges in the graph where:
 \begin{itemize}
  \item $CW \subseteq \{(c,w) \mid c,w \in V \wedge c \in C \wedge w \in W \}$ is the set of input edges, i.e., edges
  connecting input concepts to their services.
  \item $WC \subseteq \{(w,c) \mid w,c \in V \wedge w \in W \wedge c \in C \}$ is the set of output edges, i.e., edges
  connecting services with their output concepts.
  \item $CC \subseteq \{(c,c') \mid c,c' \in V \wedge c,c' \in C \wedge cmatch(c,c') \}$ is the set of edges that represent a semantic match
  between concepts.
 \end{itemize}
\end{itemize}

\begin{figure}[htpb]
  \centering
  \includegraphics[width=0.47\textwidth]{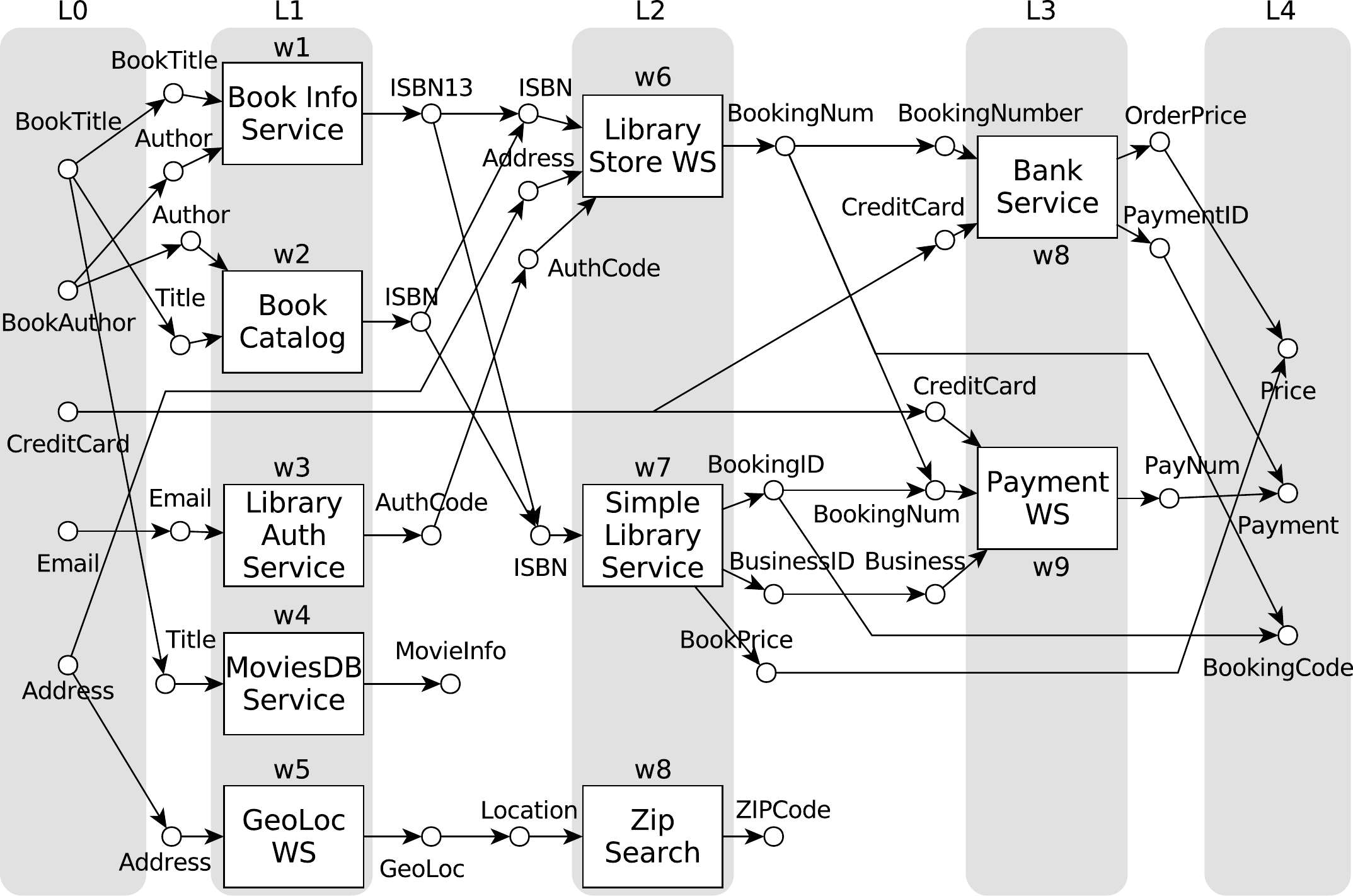}
  \caption{Composition graph example.}
  \label{Figure:GraphExample}
\end{figure}

This graph contains all the known services that could directly or indirectly be invoked given the provided inputs. The graph is divided into $N$ layers, whereby each layer $i$ has all those services whose inputs are matched by the outputs produced in previous layers and, therefore, are invokable at layer $i$. The graph is augmented with two layers, namely $L_0$ and
$L_{N+1}$. $L_0$ contains the dummy service $w_O=\{O_R, \emptyset\}$ whereas $L_{N+1}$ contains the dummy service $w_I=\{\emptyset, I_R\}$.
The first one is a service that provides as outputs the inputs of the request ($I_R$) and the last one has the goal outputs ($O_R$) as inputs.
An example of a graph for $I_R$=\textit{\{BookTitle, BookAuthor, CreditCard, Email, Address\}}
and $O_R$=\textit{\{Price, Payment, BookingCode\}} is shown in Fig. \ref{Figure:GraphExample}. 

The first step of the composition graph construction is the calculation of the relevant services.
These services can be easily calculated forwards, layer by layer, using the discovery mechanism previously presented.
Fig. \ref{Alg:FwdGraph} shows an implementation of the forward composition graph generation algorithm for a request $R$.
The algorithm selects all those services from the set of all available services $W$
that are input-relevant for the available concepts ($availCon$) in each layer using 
the $relevantIO$ function (L. 8). Then, for each \textit{input-relevant} service, the algorithm 
performs a match between the available concepts and the unmatched inputs
of each service. All the inputs that are matched are removed from the unmatched set of inputs for the current service. If there are no
unmatched inputs, then the service is invokable and thus is eligible for the current layer. 
For example, the first eligible services for the request shown in Fig. \ref{Figure:GraphExample} are the services in the layer $L1$,
which correspond with the services whose inputs are fully matched by $I_R$ (the set of concepts in $L0$). The second eligible
services are those services (placed in $L2$) whose inputs are fully matched by the outputs of the previous layers, and so on.
Note that instead of performing the invokability
check by finding a full match between $C$ and the inputs of each service, we save those inputs of each service that have been
matched before, and hence we only perform the match between the new outputs generated in the previous level ($availCon$)
and the remaining unmatched inputs of each service ($U_{set}$).
Hence, the unmatched inputs $U_{set}$ of each service decreases monotonically with each level (i.e.,
the unmatched inputs of each service always decrease when a new match is found, and the effect is propagated at each layer).
The complexity analysis for this algorithm (neglecting the optimisation effect due to the propagation of the matched inputs for simplification purposes)
is $O(l \cdot w \cdot m \cdot n + l \cdot \frac{w}{k} \cdot m \cdot n)$ which can be simplified to $O(l \cdot m \cdot n (\frac{(k+1)w}{k})$.
The first part corresponds with the complexity of the
calls to the $relevantIO$ function which is invoked $l$ times (one call per layer), whereas the second part corresponds with the complexity of the \textit{for} loop
to check the invokability of each \textit{input-relevant} service. We can expect that only a small subset of the repository $W$ is
relevant for the $availCon$ generated in the previous layer. Thus, each call to $relevantIO$ function returns a small set of relevant services
$w/k$ where $k$ ($k \gg 1$)  is a reduction factor that depends on the number of relevant services for a given set of concepts. 
This $k$ factor is different for each request and service registry. For example, if we
assume $k=100$ for a given problem for a service registry of 1,000 services, then it means that each invokation of $relevantIO(availCon, W, In)$ will 
return only the 1\% of the services of the repository ($w/k=10$). Consider the following example of a composition over a repository with
1,000 services ($w=1,000$), assuming that there are $m=5$ new output concepts generated and $n=5$ unmatched concepts at each layer, the composition
graph has 10 layers ($l=10$) and in each layer the $relevantIO$ function returns on average $w/k=10$ services (that is, $k=100$).
The complexity in this example is $10 \cdot 1000 \cdot 5 \cdot 5$ for the first part plus $10 \cdot \frac{1000}{100} \cdot 5 \cdot 5$ for the second part, which 
is $\approx 2.5 \cdot 10^5$ calls to the matchmaking system to compute all the required matches at the concept level. 

\begin{figure}
{\footnotesize
\begin{algorithmic}[1]
\Function{FwdGraph}{$R=\{I_R, O_R\}, W$}    
  \State $C := I_R;\ i := 0;\ L_0 := \{w_I\};\ L := L_0$
  \State $unmatchedIn := [\ ];\ availCon := I_R$
  \State $W' := W;$
  \Repeat
      \State $i := i + 1$
      \State $L_i := \emptyset;\ W_{selected} = \emptyset$
      \State $W_{relevant} := relevantIO(availCon, W', In)$
      \State $availCon := \emptyset$
      \ForAll{$w_i =\{I_{w_i},O_{w_i}\} \in W_{relevant}$}
	  \State $U_{set} := unmatchedIn[w_i]$
	  \State $M_{set} := Match(availCon,U_{set})$
	  \State $unmatchedIn[w_i] := U_{set} \setminus M_{set}$ 
	  \If{$M_{set} = \emptyset \wedge w_i \notin L$}
	    \State $W_{selected} = W_{selected} \cup w_i$
	    \State $availCon := availCon \cup O_{w_i}$
	  \EndIf
      \EndFor
      \State $L_i := L_i \cup W_{selected}$
      \State $W' := W' \setminus W_{selected}$
      \State $C := C \cup availCon$
  \Until{$(Match(C, O_R) = O_R) \vee L_i=\emptyset$}
  \State $L := L \cup \{w_O\}$
\EndFunction
\end{algorithmic}}
\caption{Algorithm for forward graph generation.}\label{Alg:FwdGraph}
\end{figure}

\subsubsection{Index-Based Optimisations}
Although these improvements can save search time, one of the bottlenecks of the graph generation is still 
the size of the repository $w$, which is usually some orders of magnitude bigger than the other parameters involved 
in the complexity. One effective way to reduce the impact of the
size of the repository is precalculating and indexing the input-relevant set of services for each concept of the ontology. 
The indexing of services can be done independently of any composition request as it only depends on the information available, such as the services themselves and the ontologies.

The construction of an inverted index function to recover input-relevant services or output-relevant services can be done
easily using the $relevantIO$ function. The main idea behind the inverted index is to build a key-value hash map where the
keys are the concepts of the ontology and the values are those services that are input-relevant (or output-relevant)
for that concept. This map allows to discovery relevant services in constant time during the graph generation.

We define a new function $relevantIO'$ which is the cached-version of the original function. Instead of computing the
relevance by using directly the matchmaking system, it first checks if the concept is cached in the inverted index. If
the concept is in the index, then it is immediately returned (constant time). If not, the call is delegated to the 
$relevantIO$ function. Assuming there is enough memory to keep the entire index, the index allows to
provide relevant services at $O(1)$ for each concept during the forward graph generation. 
Thus, we reduce the complexity associated to the parameter $w$. Concretely, since we can obtain at constant time
the input-relevant services for each concept, the complexity of $relevantIO(availCon, W, In)$ now depends only on
the number of concepts in $availCon$ (one access to the index per concept). Having $m=|availCon|$ (number of new concepts
at each layer) the complexity using indexes is $O(l \cdot m + l \cdot \frac{w}{k} \cdot m \cdot n)$, simplified to
$O(l \cdot m(1+\frac{w}{k} \cdot n))$.  The use of indexes to discover relevant services during the forward graph generation
has a high impact on the global performance. Using the same example as before, with $w=1000$, $l=10$, $m=5$, $n=5$ and
$k=100$ we have $10 \cdot 5 (1 + \frac{1000}{100} \cdot 5) = 2.55 \cdot 10^3$, 2 orders of magnitude lower than the
non-indexed version. 

\subsection{Graph-Based Optimisations}
Once the graph is generated, the next step is to apply different optimisations to reduce the graph size in order to
improve the optimal composition search performance. This part of the composition is independent of the discovery phase.
All the information required to search for the optimal composition is in the graph, namely, the relevant services
and the semantic relations between their inputs and outputs, so there is no need to communicate with the discovery/matchmaking
systems. We distinguish at least two different techniques \cite{RodriguezMier2011,RodriguezMier2012}: 
\textit{backward pruning} and \textit{interface dominance}.

\subsubsection{Backward pruning}
As explained earlier, the generation of the composition graph with the relevant services is done forwards, layer by layer. During this forward expansion of the graph, we are not interested in invoking services that have no explicit effects on the composition, that is,
services that are not contributing to the output goals. When the graph is completed and 
the goal outputs are reached, a backward pruning is performed to remove all non-contributing services. 
A non-contributing service is essentially a service that is not contained in the \textit{transitive closure set} of the
\textit{output-relevant} services. A service $w'=\{In_{w'},Out_{w'}\}$ is \textit{output-relevant} for a service $w=\{In_w, Out_w\}$ if
$Out_{w'} \otimes In_w \neq \emptyset$ (def. \ref{def:relevance}). Thus, the set of all \textit{output-relevant} services for
a service $w$ can be defined as:

\begin{equation}
 X(w)=\{w' \in W \mid Out_{w'} \otimes In_{w} \neq \emptyset \} 
\end{equation}

Recursively, we can define the set of $X^2(w)=X(X(w))$ as the set of \textit{output-relevant} services at the distance two. 
Extending this, the \textit{transitive closure} of the \textit{output-relevant services} can be
defined as: 

\begin{equation}
 \hat{X}(w) =X(w) \cup X^2(w) \cup X^3(w) \cup \cdots
\end{equation}

Therefore, we can say that all those services of the graph that are not in the \textit{transitive closure} of the \textit{output-relevant} services $\hat{X}$
are not contributing to the composition goals, directly nor indirectly, and can therefore be removed from the graph.

An example of this can be seen in Fig. \ref{Figure:GraphExample}. Starting from the last layer, we compute the transitive closure of the service
$w_O$, which is a dummy service that represents the goal outputs. The output relevant services for $w_O$ at distance one
are $X(w_O) = \{w_6, w_7, w_8, w_9\}$, since $Out_{w_6} \otimes In_{w_O} \neq \emptyset$ and the same for $w_7$, $w_8$ and $w_9$. We calculate now the output-relevant
services at distance two, which is $X(X(w_O))=X(\{w_6, w_7, w_8, w_9\})$. $X(\{w_6, w_7, w_8, w_9\})$ can be simply computed as the union of $X(w_6) \cup X(w_7) \cup X(w_8) \cup X(w_9)$
which is $\{w_1,w_2,w_3\}$. Repeating this, we finally have $\hat{X} = \{w_6,w_7,w_8,w_9\} \cup \{w_1,w_2,w_3\} \cup \{w_I\}$, where $w_I$ is the dummy service ommited in Fig. \ref{Figure:GraphExample} that provides the input concepts of the request (concepts in $L_0$).
Since $w_4, w_5, w_8 \notin \hat{X}$, these services ($w_4$=\textit{MoviesDB Service}, $w_5$=\textit{GeoLoc WS}, $w_8$=\textit{Zip Search}) are not contributing 
to the goals and can be removed from the graph. 

\subsubsection{Interface Dominance}
Another strategy to reduce the graph size is to analyse the equivalence and dominance of some services over others in terms of the interface they offer. It is very frequent to find services from different providers that offer similar services with overlapping interfaces. In scenarios like this, it is easy to end up with large composition graphs that make very hard to find optimal compositions in reasonable time. One way to attack this problem is to analyse the \textit{interface dominance} between services in order to find those that are equivalent or better than others in terms of the interface they provide. 

\begin{definition}
 Given a concept in a composition graph $G$ ($c \in G$), 
 we denote $\Phi(c)$ as a function that returns the set of \textit{output-relevant} services for concept $c$:
 \begin{equation}
   \Phi(c)=\{w=\{In_w, Out_w\} \in G \mid Out_w \otimes \{c\} = \{c\}\}
 \end{equation}
\end{definition}

For instance, $\Phi(Payment)$ in Fig. \ref{Figure:GraphExample} is $\{w_8,w_9\}$ since
$Out_{w_8} \otimes \{Payment\} =\{Payment\}$ and $Out_{w_9} \otimes \{Payment\} =\{Payment\}$, 
that is, concept $Payment$ is matched by an output from $w_8$ (\textit{PaymentID})
and for an output from $w_9$ (\textit{PayNum}).

\begin{definition}
 A service $w_i=\{In_{w_i}, Out_{w_i}\} \in G$ is \textit{input-equivalent} ($In_{w_i} \equiv In_{w_j}$)
 with respect to a service $w_j=\{In_{w_j}, Out_{w_j}\} \in G$ in the composition graph $G$ if:
 \begin{equation}
  \bigcup \limits _{c_i \in In_{w_i}}\{\Phi(c_i)\}
 = \bigcup \limits _{c_j \in In_{w_j}}\{\Phi(c_j)\}
 \end{equation}
\end{definition}

That is, the set of sets defined by the union of $\Phi(c)$ for each input concept $c$ of each service must be equal.
This definition formalises the idea of input equivalence of two services of the composition graph regarding
the relation between their inputs and the services that match those inputs. That means that two services $w_i$ and $w_j$
of the graph are input equivalent if the services that provide the inputs of both services 
are the same. 

\begin{definition}
 A service $w_i=\{In_{w_i}, Out_{w_i}\} \in G$ is \textit{input-dominant} ($In_{w_i} \succ In_{w_j}$)
 with respect to a service $w_j=\{In_{w_j}, Out_{w_j}\} \in G$ in the composition graph $G$ if:
 \begin{equation}
  \bigcup \limits _{c_i \in In_{w_i}}\{\Phi(c_i)\}
 \subset \bigcup \limits _{c_j \in In_{w_j}}\{\Phi(c_j)\}
 \end{equation}
\end{definition}

Thus, informally, a service is input-dominant if it only needs a subset of the information required 
by the dominated service to be invoked. For example, in Fig. \ref{Figure:GraphExample}, $w_7$ is input-dominant respect to $w_6$, since
$\{ \{w_1, w_2\} \} \subset \{ \{w_1,w_2\}, \{w_I\}, \{w_3\} \}$.

\begin{definition}
 Given a concept in a composition graph $G$ ($c \in G$), 
 we denote $\Psi(c)$ as the function that returns a set of input concepts in $G$ that are matched by $c$, that is, there exists an arc
 from $c$ to $c'$ in $G$.
 \begin{equation}
   \Psi(c)=\{c' \mid (c,c') \in G \}
 \end{equation}
\end{definition}

\begin{definition}
 A service $w_i=\{In_{w_i}, Out_{w_i}\} \in G$ is \textit{output-equivalent} ($Out_{w_i} \equiv Out_{w_j}$) 
 respect to a service $w_j=\{In_{w_j}, Out_{w_j}\} \in G$ in the composition graph $G$ if:
 \begin{equation}
  \bigcup \limits _{c_i \in Out_{w_i}}\Psi(c_i)
 = \bigcup \limits _{c_j \in Out_{w_j}}\Psi(c_j)
 \end{equation}
\end{definition}

That is, two services are output-equivalent if their outputs are matched to the same input concepts in the graph, which means that
their outputs can be consumed in the same way by the same services in $G$.

\begin{definition}
 A service $w_i=\{In_{w_i}, Out_{w_i}\} \in G$ is \textit{output-dominant} ($Out_{w_i} \succ Out_{w_j}$)
 respect to a service $w_j=\{In_{w_j}, Out_{w_j}\} \in G$ if:
 \begin{equation}
 \bigcup \limits _{c_i \in Out_{w_i}}\Psi(c_i) \supset \bigcup \limits _{c_j \in Out_{w_j}}\Psi(c_j)
 \end{equation}
\end{definition}

Therefore, one service is \textit{output-dominant} with respect to another service of the graph $G$
if their outputs match the same inputs of the same services in the composition graph
but the dominant service also provides additional outputs to the same or different services.

\begin{definition}
a service $w_i=\{In_{w_i}, Out_{w_i}\}$ is \textit{interface-equivalent} to
a service $w_j=\{In_{w_j}, Out_{w_j}\}$ ($w_i \equiv w_j$) if $In_{w_i} \equiv In_{w_j}$ and $Out_{w_i} \equiv Out_{w_j}$,
that is, both are \textit{input-equivalent} and \textit{output-equivalent}.
\end{definition}

\begin{definition}
 A service $w_i$ \textit{interface-dominates} a service $w_j$ ($w_i \succeq w_j$) if the first dominates the
second in at least one aspect (input-dominant or output-dominant) and is at least equivalent in the other aspect.
Formally, $w_i \succeq w_j$ if 
$ (In_{w_i} \succ In_{w_j} \wedge Out_{w_i} \succ Out_{w_j}) \vee
  (In_{w_i} \equiv In_{w_j} \wedge Out_{w_i} \succ Out_{w_j}) \vee
  (In_{w_i} \succ In_{w_j} \wedge Out_{w_i} \equiv Out_{w_j})$.
\end{definition}

This dominance definition can be generalised to include more features, such as preconditions, effects, or non-functional properties 
like QoS: 
\begin{definition}
A service with multiple properties $w_i=\{P^1_{w_i},P^2_{w_i},\dots,P^n_{w_i}\}$ where $P^1_{w_i}$ are the
inputs, $P^2_{w_i}$ the outputs and the rest of parameters are different properties,
dominates another service $w_j$ ($w_i \succeq w_j$) with parameters $P_{w_j}=\{P^1_{w_j}, P^2_{w_j},\dots, P^n_{w_j}\}$,
if $\forall\ k \in \{1,...,n\}\ P^k_{w_i} \succeq P^k_{w_j} \wedge \exists\ k \in \{1,...,n\}, P^k_{w_i} \succ P^k_{w_j}$.
\end{definition}

The interface dominance optimisation allows to reduce the size of the composition graph by substituting the original services of the graph by abstract interfaces that capture the functionality of the dominant or equivalent services. By minimising
the graph size we improve the performance of the search algorithms since they only explore a reduced search space. Once the search
is performed and the optimal composition workflow is generated, a post-processing step can be used to replace the abstract service
interfaces with specific implementations using the original dominant / equivalent services or by combinations of dominated services that
satisfy the same functionality of the dominant service. 

\subsection{Optimal Composition Search}
The previous optimisations are intended to reduce the composition graph but keeping the same functionality.
The next step is to perform a search over the graph to find the best composition among all the possible compositions
that satisfy the input/output request. The search can be designed to optimise different criteria, such as the number
of services, the execution path length or QoS properties. Typically, the search over the graph can be done forwards or backwards. 
In the first case, the composition starts from the inputs of the request (first layer), selecting invokable 
services until the goal outputs are obtained, whereas the second case starts with the goal outputs (last layer), 
selecting relevant services for the outputs until a composition that can be invoked with the initial inputs is found.

Formally, the composition search can be modelled as a state-transition system, where the problem is divided into a set
of states and transitions between states \cite{ghallab2004automated}. A state transition system is defined
as a 3-tuple $\Sigma = (S,A,\gamma)$, where:

\begin{itemize}
 \item $S = \{s_1,s_2,\dots\}$ is a finite set of states.
 \item $A = \{a_1,a_2,\dots\}$ is a finite set of actions.
 \item $\gamma : S \times A \rightarrow S$ is a state-transition function.
\end{itemize}

Using the concept of the state-transition system, the state space search problem can be defined as $P=\{\Sigma, s_0,  G\}$, where $s_0 \in S$ is the
initial state and $G \subseteq S$ is a set of goal states.

The state-transition system $\Sigma$ allows the search to navigate through the set of states applying different actions,
where each action may be associated to a cost that we want to minimise. The state representation may vary depending on the
strategy used. Typically, in the case of the backward search, the state will contain the information of the unsatisfied concepts
at each state, starting with the goal outputs. 
The goal then is to find a succession of actions $\langle a_1, a_2, \dots, a_n \rangle$ with the minimum cost that leads
from the initial state, where unsatisfied concepts = goal outputs, to the goal state, where unsatisfied concepts = $\emptyset$, that is, there are no unsatisfied
concepts and the composition is invokable. The available transitions between states are given by the applicable actions to each state, i.e.,
the output relevant services that can be selected to resolve all the unsatisfied concepts.

Given a composition graph $G=(V,E)$ as defined previously, where $V=W \cup C$ is the set of vertices which are the services 
and the concepts (inputs/outputs) of the graph, the state-transition system $\Sigma$ for the (backward) composition problem is defined as follows:
\begin{itemize}
 \item $S \subseteq 2^{|C|}$ where $C$ is the set of all concepts in the composition graph, i.e., a state is a set of concepts of the graph, $s=\{c_1,\dots,c_n\}$.
 \item $A \subseteq 2^{|W|}$ where $W$ is the set of services in the composition graph, i.e., an action is a set of services from the graph, $a=\{w_1,\dots,w_n\}$.
 \item $\gamma(a,s) = (s - \bigcup (\Psi(c_i) \mid c_i \in Out(a)) \cup In(a))$, i.e., the application of an action $a=\{w_1,\dots,w_n\}$
 to a state $s=\{c_1,\dots,c_n\}$ generates a new state where all concepts that are matched by the outputs of the services of the actions are removed, and
 the inputs of the services of the actions are added as the new unsatisfied concepts. Functions $In(a)$ and $Out(a)$ return the union of the input concepts
 and the union of the output concepts of the services in $a$ respectively. 
\end{itemize}

The initial state $s_0$ of the backward composition problem $P=(\Sigma, s_0, G)$ is defined as $s_0=In_{w_O}$, i.e., 
the input concepts of the output dummy service. For example, in Fig. \ref{Figure:GraphExample}, the initial state is $s_0=\{i_{18}, i_{19}\}$. 
On the other hand, there is just one goal state $G=\{s_g=\emptyset\}$, i.e., the goal state is reached when there are no unsatisfied concepts in the composition.

The efficiency of the search can also be improved using \textit{search optimisations} depending on the search strategy followed. 
These optimisations can be applied to the available actions for each state by pruning actions that lead to dead-ends, actions that are equivalent,
or actions that are dominated (cannot lead to a better solution).

\section{Reference Implementation} 
\label{Section:Architecture}

\begin{figure*}[htpb]
  \centering
  \includegraphics[width=0.90\textwidth,height=\textheight,keepaspectratio]{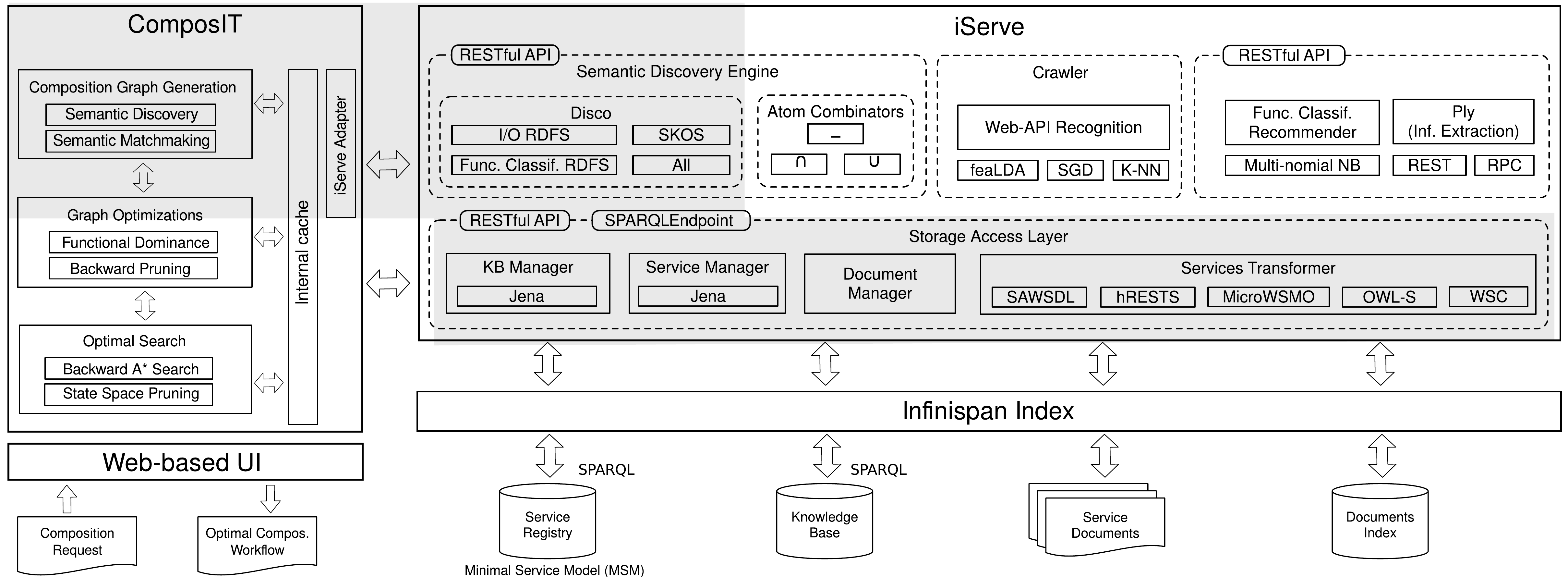}
  \caption{ComposIT / iServe architecture}
  \label{Figure:Architecture}
\end{figure*}

We developed a reference implementation of 
the integrated graph-based composition framework that is based on two main components:
iServe \cite{Pedrinaci2010}, a service warehouse with advanced discovery support which provides the service 
registry and takes care of the matchmaking and service discovery activities, 
and ComposIT \cite{RodriguezMier2012}, which is in charge of the graph-based composition part.

Fig. \ref{Figure:Architecture} depicts the architecture of the system. In a nutshell the composition process is carried out as follows. When a composition request is sent to the system through the Web UI, ComposIT starts computing the composition graph with all the relevant services for the request. To this end, all the relevant services are discovered layer by layer using the fine-grained I/O logic-based discovery support provided by the Semantic Discovery Engine of iServe. This engine relies on the \textit{Service Manager} and the \textit{KB Manager} to retrieve the relevant services using semantic reasoning capabilities. During the composition graph generation, ComposIT also makes intensive use of the \textit{KB Manager} in order to carry out concept level matching and consequently figure out how the inputs and outputs 
of the services obtained can be connected. Once the composition graph is generated, ComposIT applies the \textit{backward pruning} and the \textit{interface dominance} optimisations to reduce the graph size. These
optimisations are applicable using only the information contained in the graph, and thus there is no need to interact with the discovery component. Finally, an optimal search is performed over the graph using a backward A* algorithm that extracts the optimal composition
from the graph.

In the next sections we shall cover in more detail the inner workings of iServe and ComposIT respectively.

\subsection{iServe}
iServe~\cite{Pedrinaci2010}, see right hand-side of Fig. \ref{Figure:Architecture}, is a service warehouse whose functionality includes the core service registry anchored on Linked Data principles, semantic reasoning support, advanced discovery functionality, and further analysis components able to assist in automatically locating and generating semantic service descriptions out of Web resources. For the purposes of this work we have essentially exploited the registry and discovery functionality. 

The service discovery functionality builds on top of the Storage Access Layer, which is in charge of managing the registry’s data that includes Service descriptions, related documents and the corresponding Ontologies. This layer essentially provides a RDF/S and OWL storage and reasoning support, document storage, as well as basic crawling facilities to automatically obtain referenced Ontologies. RDF/S and OWL storage and reasoning support is delegated to dedicated engines which are accessed by means of the SPARQL 1.1 standard. Therefore, the reasoning capabilities depend largely on the actual configuration of the store.
Concretely, the discovery infrastructure contacts the Service Manager to list services given basic criteria such as the input and output types provided, and the KB Manager to obtain concepts, properties, and their sub or super concepts. Depending on their implementation Service and KB Managers combine internal indexes with \textit{SPARQL} queries issued to the triple store by means of Jena. 

Services are imported to iServe using a range of transformation engines able to import service descriptions in a variety of formalisms including SAWSDL, WSMO-Lite, OWL-S, and MicroWSMO. These plugins generate descriptions expressed in terms of a simple RDF/S model,
Minimal Service Model (MSM) \cite{Pedrinaci2010}, which essentially captures the intersection of existing service description formalisms.
By means of these transformations iServe provides an homogeneous description for services that were orginally annotated using heterogeneous means.

Given that, as we saw in Section~\ref{Section:Framework}, the response time of the overall composition is highly dependent on the performance of the service discovery and concept matchmaking tasks, we extended iServe with various implementations of the Service and Knowledge Base Managers. We tested different configurations to study their individual performance and the overall impact on composition response times. In particular, we used the following configurations:

\begin{enumerate}
  \item \textit{SPARQL D/M}: pure \textit{SPARQL} Discovery / Matchmaking where all interactions with the Service and Knowledge Base managers are directly implemented as \textit{SPARQL} queries. This is the typical approach of discovery engines and was the original implementation of iServe. 
  \item \textit{Index. D/SPARQL+Cache M}: I/O service discovery is based on an index. We additionally used herein an intermediate cache at the level of the concept matcher in order to avoid issuing recurrent SPARQL queries.  
  \item \textit{Full Indexed D/M}: both service discovery and concept matchmaking relied on local indexes pre-populated at load time (and updated with writes). In this configuration, service discovery and concept matchmaking do not need to issue any SPARQL query to the backed.
\end{enumerate}

\subsection{ComposIT}
ComposIT~\cite{RodriguezMier2012}, depicted in the left hand-side of Fig. \ref{Figure:Architecture}, is the semantic Web service composition engine we rely on. 
It implements all the different graph-based composition phases of the framework described in Sec. \ref{Section:Framework}. 
The semantic service discovery and matchmaking mechanisms, which originally were directly implemented internally, are delegated to iServe by means of 
integration adapters implemented for the purposes of this work. ComposIT nonetheless uses an internal cache and an index to efficiently recover the 
information of the generated composition graph. 
It is worth to note that the architecture supports the deployment of multiple, distributed iServe instances
to provide different endpoints that can be used by ComposIT in the composition phase by aggregating the results of the registries at the ComposIT API level.
Indeed, since the services to contemplate at composition time are identified by the remote registry and we just use them directly, composing this set of services out of just one API call or several calls in parallel (one per registry) is a trivial change. The overall response time analysis would still remain unchanged, and would have an upper-bound determined by the slowest registry. This also applies to other third-party discovery engines as long as they support fine-grained I/O discovery queries as described in Sec. \ref{Subsection:Discovery}. The integration of these third-party registries could be achieved by developing interface adapters (with capabilities to retrieve input and output relevant services) which could be plugged in to the system, keeping the generation of the composition graph isolated from the concrete registries used.

The generated composition graph can contain different compositions with the same or different length (number of layers) and with different number
of services depending on the services that have been selected to generate the needed data. Among the different combinations that can be obtained, 
the goal of ComposIT is to find the shortest service composition with the minimum number of services. For this purpose,
ComposIT searches for the optimal composition by carrying out a heuristic search based on the A* algorithm \cite{Hart1968a}. This search was implemented
using Hipster4j~\cite{Rodriguez-Mier2014} to identify a minimal subset of the services from the graph that satisfy the request (in terms of inputs and outputs). Note that multiple compositions can be extracted from the composition graph since there may be different services that generate outputs of the same concept.

\section{Evaluation}
\label{Section:Evaluation}

In this section we present a quantitative evaluation of our approach.
The purposes of the evaluation are: 1) measure the scalability of the approach with many services; 2) study the impact
of the discovery on the overall composition performance and 3) compare the performance with different optimisations. 

In order to perform a standard and comparable evaluation, we selected the Web Service Challenge 2008 (WSC'08) service datasets. These datasets allow us to measure the scalability with an increasingly large set of services (from 158 to 8,119 services). Services were imported to iServe using an specific transformer plugin which translates each service description in
the WSC'08 XML format into MSM, and the XML concept taxonomy into an equivalent OWL representation. iServe is responsible of identifying, loading and reasoning with the ontologies used
in the service descriptions. Data types of the input and outputs of service descriptions are linked to their corresponding
semantic concepts through the $modelReference$ property of the MSM, which points to the concepts defined in the transformed OWL model. 

Experiments were run under Ubuntu 10.04 64-bit on a PC with an Intel Core 2 Duo E6550 at 2.33GHz and 4 GB of RAM. OWLIM-Lite 5.3 with OWL Horst reasoning was chosen in iServe as the RDF triple store for the semantic registries and deployed within Tomcat 7.

\begin{table}[htpb]
\caption{Characteristics of the WSC'08 datasets.}
\centering
\scalebox{0.8}{
\begin{tabular}{|c|c|c|c|c|c|c|c|}
\hline
\textbf{Dataset} & \textbf{\#Serv.} & \textbf{\#Con.} & \textbf{\#Serv.Sol.} & \textbf{Length}  \\ \hline
WSC'08 01 & 158 & 1,540 & 10 & 3   \\ \hline
WSC'08 02 & 558 & 1,565 & 5 & 3  \\ \hline
WSC'08 03 & 604 & 3,089 & 40 & 23 \\ \hline
WSC'08 04 & 1,041 & 3,135 & 10 & 5 \\ \hline
WSC'08 05 & 1,090 & 3,067 & 20 & 8 \\ \hline
WSC'08 06 & 2,198 & 12,468 & 40 & 9 \\ \hline
WSC'08 07 & 4,113 & 3,075 & 20 & 12 \\ \hline
WSC'08 08 & 8,119 & 12,337 & 30 & 20 \\ \hline
\end{tabular}
}
\label{Table:WSC}
\end{table}

Table \ref{Table:WSC} shows the characteristics of each WSC'08 dataset. The number of services and concepts in the ontology
of each dataset are shown in columns \textit{\#Serv.} and \textit{\#Con.} respectively. The quality of the solutions is based on the number of services and the length (i.e., number of layers) of the composition. The optimal quality of solution for each dataset (according to the WSC'08 competition) are shown in columns \textit{\#Serv.Sol.} and \textit{Length}.

Experimentation was done using the configurations explained in Sec. \ref{Section:Architecture} with one instance of iServe 
in order to measure the effect of the Discovery/Matchmaking over the whole composition process.
Results with each configuration are shown in Table \ref{Table:Results}.
The second column shows the size (number of services) of the resulting composition graph for each dataset. The next columns show the time taken to generate the composition graph (\textit{G. time}) in seconds and the number of \textit{SPARQL} queries generated during that process. The last three columns show the size of the graph after the graph-based optimisations, the time of the composition search (graph optimisations + optimal A* backward search) and the number of services and length of the optimal composition found. Note that the backward optimal search does not depend on the configuration 
selected since it only uses the information in the composition graph.

\begin{figure}[htpb]
  \centering
  \includegraphics[width=0.40\textwidth,height=\textheight,keepaspectratio]{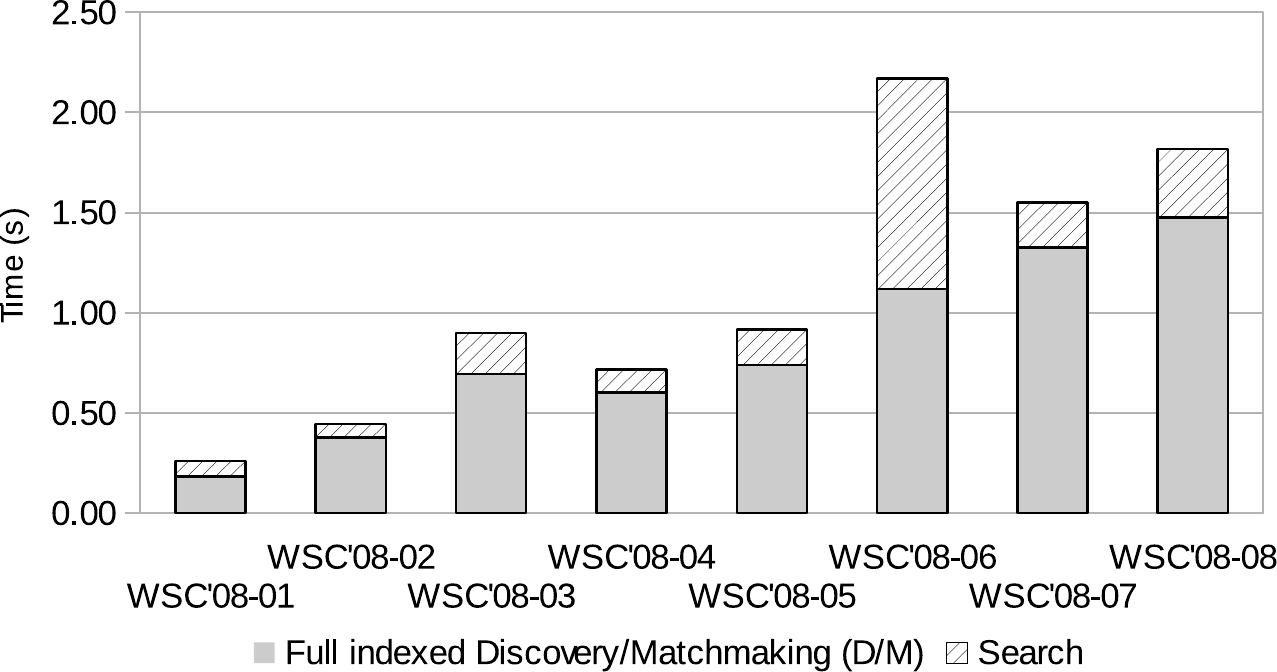}
  \caption{Graph generation time vs Search time for the Full Indexed Discovery/Matchmaking configuration.}
  \label{Figure:DiscoveryCompositionImpact}
\end{figure}

\begin{table*}[htbp]
\caption{Evaluation results with different Discovery/Matchmaking (D/M) configurations with the WSC'08 datasets}
\scalebox{0.82}{
\begin{tabular}{cc|c|c|c|c|c|c|ccc}
\cline{3-10}
 & & \multicolumn{6}{|c|}{\textbf{Discovery/Matchmaking (D/M)}} & \multicolumn{2}{|c|}{\textbf{Composition}} & \\
\cline{3-10}
 & & \multicolumn{2}{|c|}{\textbf{1) SPARQL D/M}} & \multicolumn{2}{|c|}{\textbf{2) Index. D/SPARQL+Cache M}} & \multicolumn{2}{|c|}{\textbf{3) Full Indexed D/M}} & \multicolumn{2}{|c|}{ } &  \\ \hline
\multicolumn{1}{|c|}{ \textbf{Dataset} } & \multicolumn{1}{|c|}{\textbf{G. size}} & \textbf{G. time (s)} & \textbf{\#SPARQL} & \textbf{G. time (s)} & \textbf{\#SPARQL} & \textbf{G. time (s)} & \textbf{\#SPARQL} & \textbf{G. size (opt)} & \multicolumn{1}{|c|}{\textbf{Comp. time (s)}} & \multicolumn{1}{|c|}{\textbf{Sol. (serv./length)}} \\ \hline
\multicolumn{1}{|c|}{WSC'08-01} & \multicolumn{1}{|c|}{35}   & 28.52   & 3256   & 5.67  & 624   & 0,18 & 0 & 13 (-37\%) & \multicolumn{1}{|c|}{0.08} & \multicolumn{1}{|c|}{10/5} \\ \hline
\multicolumn{1}{|c|}{WSC'08-02} & \multicolumn{1}{|c|}{35}   & 63.30   & 7349   & 11.76 & 1830  & 0,38 & 0 & 13 (-37\%)&  \multicolumn{1}{|c|}{0.07} & \multicolumn{1}{|c|}{5/3} \\ \hline
\multicolumn{1}{|c|}{WSC'08-03} & \multicolumn{1}{|c|}{105}  & 262.80  & 36619  & 20.05 & 3184  & 0.69 & 0 & 40 (-38\%)&  \multicolumn{1}{|c|}{0.21} & \multicolumn{1}{|c|}{40/23} \\ \hline
\multicolumn{1}{|c|}{WSC'08-04} & \multicolumn{1}{|c|}{44}   & 136.20  & 13828  & 21.12 & 3481  & 0.60 & 0 & 25 (-57\%)&  \multicolumn{1}{|c|}{0.12} & \multicolumn{1}{|c|}{10/5} \\ \hline
\multicolumn{1}{|c|}{WSC'08-05} & \multicolumn{1}{|c|}{97}   & 333.60  & 41148  & 26.05 & 4417  & 0.74 & 0 & 52 (-54\%)&  \multicolumn{1}{|c|}{0.18} & \multicolumn{1}{|c|}{20/8} \\ \hline
\multicolumn{1}{|c|}{WSC'08-06} & \multicolumn{1}{|c|}{189}  & 1051.20 & 93682  & 48.21 & 8511  & 1.12 & 0 & 75 (-40\%)&  \multicolumn{1}{|c|}{1.05} & \multicolumn{1}{|c|}{42/7} \\ \hline
\multicolumn{1}{|c|}{WSC'08-07} & \multicolumn{1}{|c|}{124}  & 1183.20 & 120881 & 35.76 & 6376  & 1.33 & 0 & 70 (-56\%)&  \multicolumn{1}{|c|}{0.23} & \multicolumn{1}{|c|}{20/12} \\ \hline
\multicolumn{1}{|c|}{WSC'08-08} & \multicolumn{1}{|c|}{121}  & 1656.00 & 89518  & 78.00 & 15844 & 1.48 & 0 & 58 (-48\%)&  \multicolumn{1}{|c|}{0.34} & \multicolumn{1}{|c|}{30/20} \\ \hline
\end{tabular}
}
\label{Table:Results}
\end{table*}

The analysis of these results reveals that the discovery and matchmaking phases take most of the time of the composition,
even using the optimal configuration (\textit{Full Indexed D/M}) to avoid the latency of the \textit{SPARQL} queries. 
This is graphically represented in Fig. \ref{Figure:DiscoveryCompositionImpact}. This figure shows the overall composition time for each dataset including the relative time of the \textit{Full Indexed D/M} (blue bar) and the \textit{Composition Search} (red bar). The \textit{Full Indexed D/M} takes 77\% of the total composition time on average. This percentage is even higher (about 99\%) if the discovery and matchmaking are not optimised using indexes and cache. In other words, as anticipated by the complexity analysis presented earlier, discovery and matchmaking are responsible for the majority of the computation that needs to be performed to compose services. Optimising both phases is thus fundamental.

\begin{figure}[htpb]
  \centering
  \includegraphics[width=0.45\textwidth,height=\textheight,keepaspectratio]{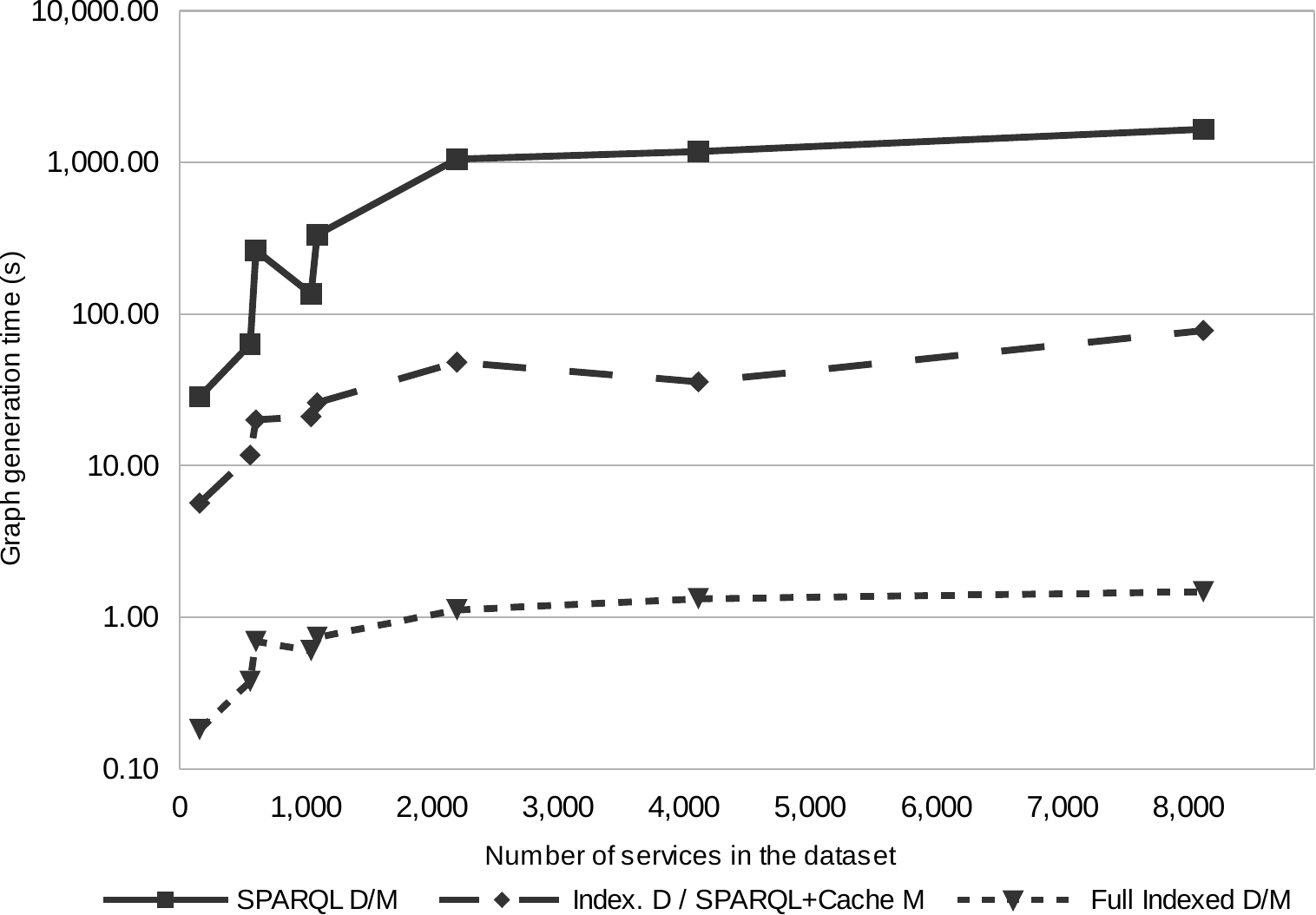}
  \caption{Composition time for different configurations.}
  \label{Figure:Comparison}
\end{figure}

The comparison of the scalability of the three configurations with respect to the number of services is shown in Fig. \ref{Figure:Comparison}. As can be seen, directly querying the backend (see \textit{SPARQL} D/M), which is the approach followed by most discovery engines, rapidly becomes prohibitively slow taking 1,656 seconds (i.e., 27.6 min) in the largest dataset. Indeed, the generation of the composition graph requires computing every semantic match between all inputs and outputs as well as discovering relevant services at each layer. Doing so leads to issuing thousands of \textit{SPARQL} queries. This can be dramatically improved using a discovery index and a local cache for the matchmaking system as can be seen in the second configuration. In this case, almost every composition is calculated in less than
a minute. The generated \textit{SPARQL} queries in this case are reduced by up to 91\% (for the WSC'08-3 dataset) leading to a significant performance improvement. Although such an improvement can be enough to solve the smaller datasets in a few
seconds, the latency of the \textit{SPARQL} queries still remains a bottleneck for bigger datasets like the WSC'08-08 dataset that still require evaluating 15,844 \textit{SPARQL} queries for generating the composition graph in 78 seconds. Our tests show, however, that the full indexed configuration allows solving the largest problems very fast by avoiding the evaluation of \textit{SPARQL} queries at composition time. This configuration entails the derived need for service registries to additionally calculate and maintain the indexes. Doing so, nonetheless, enables performing very efficient composition over remote 3rd party controlled service registries akin to what can be obtained by the fastest composition engines in the unrealistic scenarios where all services are available and pre-loaded in memory. Additionally, indeed, using those indexes allows service registries to provide highly efficient discovery for a controlled set of queries, while retaining the ability to offer fully flexible yet less efficient 
discovery support. 

We have also evaluated our framework with the WSC'09-10 datasets. Results show a similar scalability behaviour with the number of services for each configuration.
Moreover, our approach is able to solve all the datasets with optimal results, which are shown at https://wiki.citius.usc.es/composit:wsc09.

\section{Conclusions}
\label{Section:Conclusions}
In this paper we have presented a theoretical analysis of service composition in terms of its dependency with service discovery. Driven by this analysis we have defined a formal integrated graph-based composition framework anchored on the integration of service discovery and matchmaking within the composition process. We have devised a reference implementation of this framework on the basis of two pre-existing separate components, namely iServe and ComposIT. This reference implementation has been used to empirically study the impact of discovery and matchmaking on service composition, and we have provided three different configurations with varying performance. Our empirical analysis shows that, indeed, typical approaches followed by discovery engines cannot serve as a suitable basis to support efficient service composition as they lead to prohibitive execution times. We have also shown, though, that with the adequate interface granularity and indexing, discovery engines can support highly efficient 
composition akin to that obtained by the fastest composition engines without having to assume to local availability and in-memory preloading of service registries. 

This work proves the scalability and flexibility of our proposal and provides insights on how integrated composition systems can be designed in order to achieve good performance in real scenarios, where service registries and composition frameworks are likely to be distributed and controlled by diverse organisations.

\section*{Acknowledgment}
This work was partly supported by the Spanish Ministry of Economy and Competitiveness (MEC) under grant TIN2011-22935, and by the COMPOSE European Project (FP7-ICT-317862). 
Pablo Rodríguez-Mier is supported by an FPU Grant from the MEC (ref. AP2010-1078) and was also partially funded by Pedro Barrié de la Maza Foundation (2013).

\bibliographystyle{IEEEtran}
\bibliography{bibliography}

\begin{IEEEbiographynophoto}{Pablo Rodríguez-Mier}
is a PhD student at CiTIUS, Universidade de Santiago de Compostela, Spain. His research interests include
heuristic search and automatic Web service composition. 
\end{IEEEbiographynophoto}

\vskip -10mm

\begin{IEEEbiographynophoto}{Dr. Carlos Pedrinaci}
Dr. Carlos Pedrinaci is a Research Fellow of the Knowledge Media Institute at The Open University, UK. He holds a PhD in Artificial Intelligence from the University of the Basque Country (Spain). His research interests include Web Science, Semantic Web,  and Service Science. Carlos has been actively involved in the standardization of Semantic Web Services technologies as part of the W3C SAWSDL Working Group and, recently, the Linked USDL initiative. 
\end{IEEEbiographynophoto}

\vskip -10mm

\begin{IEEEbiographynophoto}{Dr. Manuel Lama}
is Associate Professor at CiTIUS, Universidade de Santiago de Compostela. His research interests focuses on discovery and
composition of Web services, semantic annotation, and process mining.
\end{IEEEbiographynophoto}

\vskip -10mm

\begin{IEEEbiographynophoto}{Dr. Manuel Mucientes} 
is Associate Professor in Computer Science and Artificial Intelligence within
the CiTIUS of the Universidade de Santiago de Compostela. His current research
interests are evolutionary computation, robotics, and Web services.
\end{IEEEbiographynophoto}

\end{document}